\documentclass[lettersize,journal]{IEEEtran}
\usepackage{amsmath,amsfonts}
\usepackage{algorithmic}
\usepackage{algorithm}
\usepackage{array}
\usepackage{textcomp}
\usepackage{stfloats}
\usepackage{url}
\usepackage{verbatim}
\usepackage{graphicx}
\usepackage{epstopdf}
\usepackage{cite}
\usepackage{booktabs}
\usepackage{adjustbox}
\usepackage{multirow}
\usepackage{hyperref}
\hypersetup{hidelinks}
\usepackage{mathtools}
\usepackage{adjustbox}
\usepackage{multirow}
\usepackage{enumitem}
\usepackage{multirow}
\usepackage{diagbox}
\usepackage{bm}
\usepackage{amsmath}  
\usepackage{amssymb}  
\usepackage{xcolor}
\usepackage{url}
\usepackage{pifont}
\usepackage{hyperref}
\usepackage{cleveref}
\begin{document}

\title{Image Forgery Localization with State Space Models}

\author{Zijie Lou, 
Gang Cao,~\IEEEmembership{Member,~IEEE,}
Kun Guo,
~Shaowei Weng,~\IEEEmembership{Member,~IEEE,}
and Lifang Yu
\thanks{
 
Zijie Lou, Gang Cao and Kun Guo are with the School of Computer and Cyber Sciences, Communication University of China, Beijing 100024, China, and also with the State Key Laboratory of Media Convergence and Communication, Communication University of China, Beijing 100024, China (e-mail: \{louzijie2022, gangcao, kunguo\}@cuc.edu.cn).

Shaowei Weng is with the Fujian Provincial Key Laboratory of Big Data Mining and Applications, Fujian University of Technology, Fuzhou 350118, China (e-mail: wswweiwei@126.com).

Lifang Yu is with the Department of Information Engineering, Beijing Institute of Graphic Communication, Beijing 100026, China (e-mail: yulifang@bigc.edu.cn).

}
}

\markboth{Journal of \LaTeX\ Class Files, Vol. 14, No. 8, August 2015}
{Shell \MakeLowercase{\textit{et al.}}: Bare Demo of IEEEtran.cls for IEEE Journals}
\maketitle

\begin{abstract}
Pixel dependency modeling from tampered images is pivotal for image forgery localization.  Current approaches predominantly rely on Convolutional Neural Networks (CNNs) or Transformer-based models, which often either lack sufficient receptive fields or entail significant computational overheads.  Recently, State Space Models (SSMs), exemplified by Mamba, have emerged as a promising approach. They not only excel in modeling long-range interactions but also maintain a linear computational complexity. In this paper, we propose LoMa, a novel image forgery localization method that leverages the selective SSMs. Specifically, LoMa initially employs atrous selective scan to traverse the spatial domain and convert the tampered image into ordered patch sequences, and subsequently applies multi-directional state space modeling.  In addition, an auxiliary convolutional branch is introduced to enhance local feature extraction. Extensive experimental results validate the superiority of LoMa over CNN-based and Transformer-based state-of-the-arts. To our best knowledge, this is the first image forgery localization model constructed based on the SSM-based model. We aim to establish a baseline and provide valuable insights for the future development of more efficient and effective SSM-based forgery localization models. Code is available at \href{https://github.com/multimediaFor/LoMa}{https://github.com/multimediaFor/LoMa.}
\end{abstract}


\IEEEpeerreviewmaketitle

\section{Introduction}
Image forgery localization (IFL), which aims to segment tampered regions in an image, is a fundamental yet challenging digital forensic task. Its utility spans many practical applications, including digital forensics, copyright protection and identity verification. IFL workflows typically encompass two primary stages: first, capturing the pixel dependency of input tampered images, and secondly, leveraging such information to generate localization map of forged region. Current methods \cite{liu2022pscc, dong2022mvss, kwon2022learning, guo2024effective, guillaro2023trufor, lou2024exploring} for modeling pixel dependency predominantly rely on CNNs and Transformers-based models. However, as illustrated in Table \ref{mamba}, such methods either (1) do not have sufficient receptive fields to capture inter-pixel correlations, or (2) suffer from prohibitive computational complexity.

\begin{table}
\centering
\caption{Comparison of the model design for pixel dependency modeling of LoMa and existing methods. LoMa enjoys both the advantages of a large receptive field and linear complexity.}

\begin{adjustbox}{width=\linewidth}
\begin{tabular}{l|ccc} 
\toprule
\multirow{2}{*}{\textbf{Architecture}}  & \textbf{Linear} & \textbf{Global} & \textbf{Representative} \\ 
                 & \textbf{Complexity} & \textbf{Receptive Field} & \textbf{Method} \\ 
\midrule
CNN  & \ding{51}   & \ding{55}  &  CAT-Net \cite{kwon2022learning} \\
Transformer  & \ding{55}   & \ding{51}  &  MPC \cite{lou2024exploring} \\
Mamba & \ding{51}   & \ding{51}  & LoMa (Ours) \\
\midrule
\end{tabular}
\end{adjustbox}
\label{mamba}
\end{table}

On the other hand, Natural Language Processing (NLP) has recently seen the advent of structured state space models \cite{gu2021efficiently}. From a theoretical perspective, SSMs integrate the advantages of Recurrent Neural Networks (RNNs) with those of CNNs, capitalizing on the global receptive field characteristics of RNNs while also benefiting from the computational efficiency of CNNs. A particularly noteworthy SSM is the Selective State Space Model (S6), also known as Mamba \cite{gu2023mamba}, which has attracted considerable interest within the vision research community. The novelty of Mamba lies in its ability to render SSM parameters time-variant (i.e., dependent on the data), allowing it to effectively identify pertinent contexts within sequences—a critical factor for improving model performance. Nevertheless, to the best of our knowledge, \textbf{Mamba has not yet been applied to the image forgery localization task.}

In this paper, we propose LoMa, a novel image forgery localization method that adapts the state space model for efficient global pixel dependency modeling. As shown in Table \ref{mamba}, LoMa provides the advantages of a global receptive field with linear complexity. Specifically, our method introduces the Mixed-SSM Block, which initially utilizes atrous selective scanning to navigate the spatial domain and convert the tampered image into ordered patch sequences, followed by the implementation of multi-directional state space modeling. Moreover, an auxiliary convolutional branch is integrated to enhance the extraction of local features. After modeling the pixel dependencies with the SSM and CNN blocks, pixel-wise localization results are achieved using a lightweight MLP decoder. Extensive experimental results demonstrate that LoMa achieves superior localization accuracy compared to existing state-of-the-art CNN- and Transformer-based approaches, while simultaneously attaining computational complexity reduction through lightweight architectural innovations.

\label{sec:intro}
\setlength{\parindent}{1em}

The rest of this letter is organized as follows. The proposed LoMa scheme is described in Section II, followed by extensive experiments and discussions in Section III. We draw the conclusions in Section IV.

\section{Proposed LoMa Scheme}
\label{sec:method}

\begin{figure*}[!t]
\centering
\includegraphics[width=\textwidth]{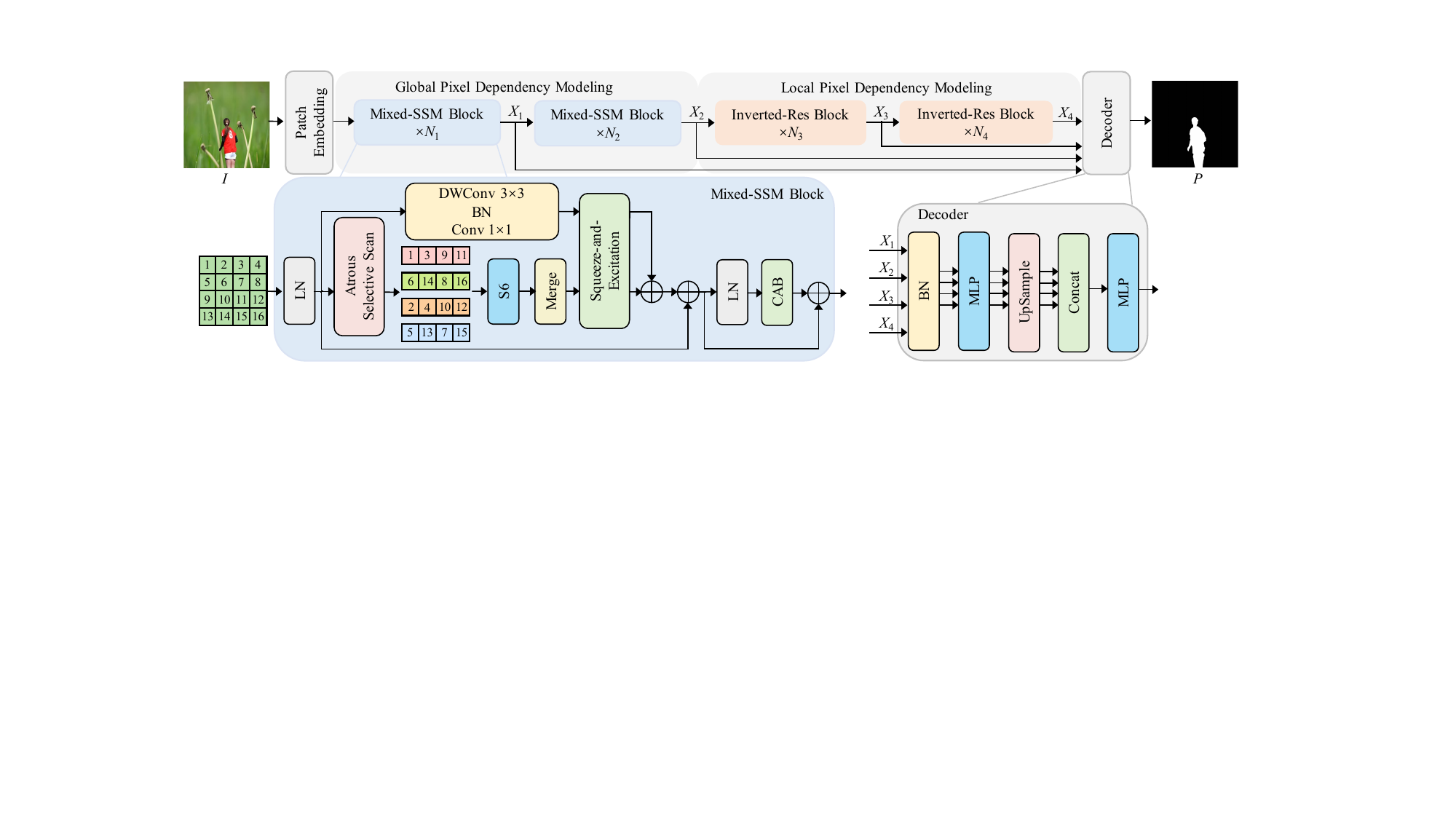}
\caption{Proposed image forgery localization scheme LoMa. $N_i$=\{2, 2, 9, 2\}, LN, BN  and CAB refer to Layer Normalization, Batch Normalization, and Channel Attention Block, respectively.}
\label{framework}
\end{figure*}

\begin{figure*}[!t]
\centering
\includegraphics[width=\textwidth]{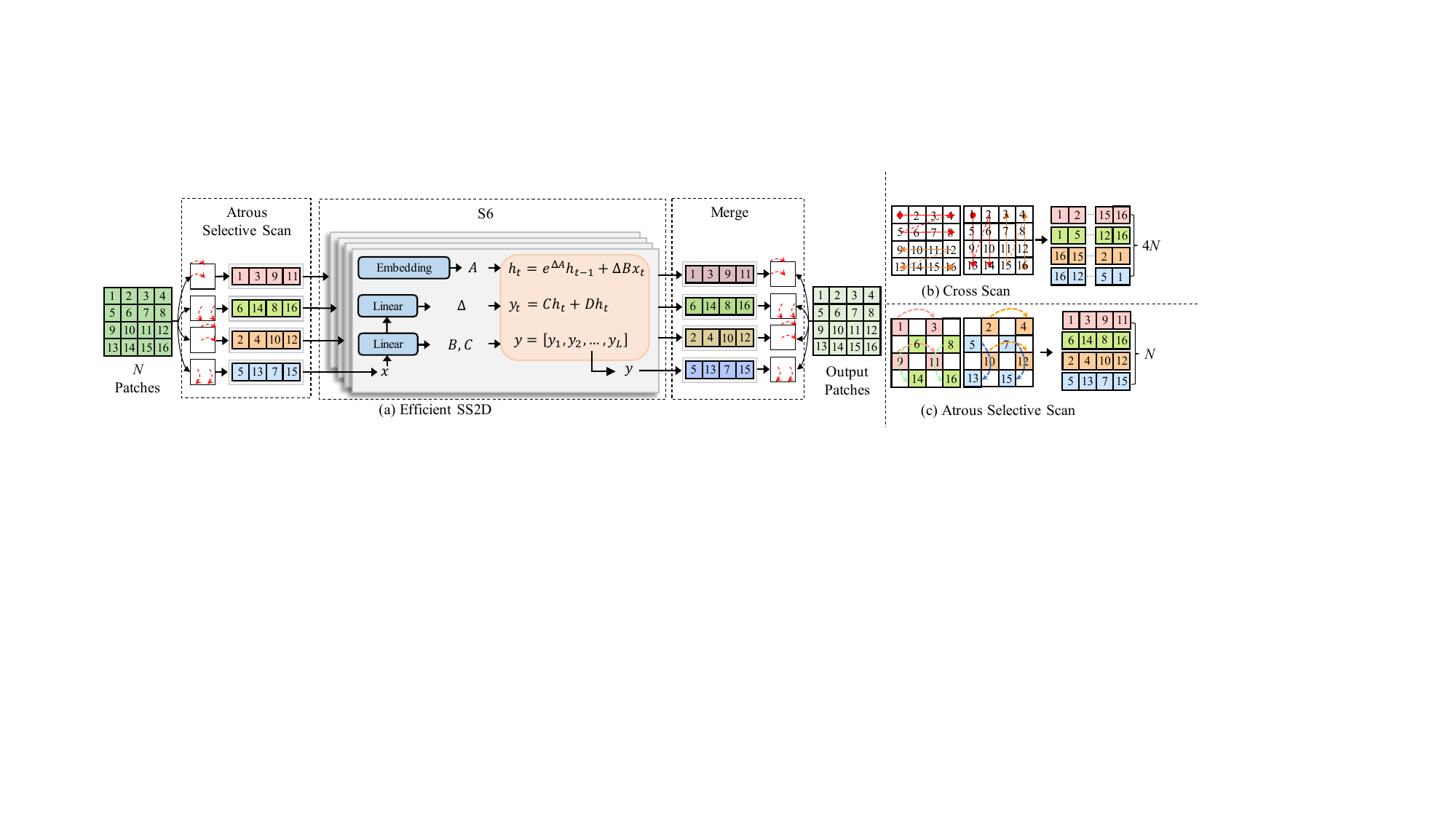}
\caption{Illustration of efficient 2D-Selective-Scan (SS2D).}
\label{ss2d}
\end{figure*}

\subsection{Preliminaries}
State Space Models are primarily motivated by continuous linear time-invariant systems, which maps a 1-dimensional function or sequence $x_{t} \in \mathbb{R} \rightarrow y_{t} \in \mathbb{R}$ through an implicit latent state $h_{t}\in \mathbb{R}^{N}$. In particular, SSMs can be formulated as an ordinary differential equation:

\begin{equation}
\begin{aligned}
\label{eq:ssm}
    h'_{t}&=Ah_{t}+Bx_{t}\\
    y_{t}&=Ch_{t}+Dx_{t}
\end{aligned}
\end{equation}
\noindent
where $N$ is the state size, $A \in \mathbb{R}^{N\times N}$, $B \in \mathbb{R}^{N \times 1}$, $C \in \mathbb{R}^{1\times N}$, and $D \in \mathbb{R}$. After that, the discretization process is typically adopted to integrate Eq. (\ref{eq:ssm}) into practical deep learning algorithms. Recent works on state space models \cite{gu2021efficiently} propose the introduction of a timescale parameter $\Delta$ to transform the continuous parameters $A$ and $B$ to discrete parameters $\overline{A}$ and $\overline{B}$, i.e.,

\begin{equation}
\begin{aligned}
    \overline{A} &= \exp\left( \Delta A \right)\\ 
    \overline{B} &= \left( \Delta A \right)^{-1} \left( \exp \left( \Delta A - I \right) \right)  \cdot \Delta B
\end{aligned}
\end{equation}

After the discretization, the discretized version of Eq. (\ref{eq:ssm}) with step size $\Delta$ can be rewritten in the following form:

\begin{equation}
\begin{aligned}
\label{eq:ssm2}
    h_{t}&=\overline{A}h_{t-1}+\overline{B}x_{t}\\
    y_{t}&=Ch_{t}+Dx_{t}
\end{aligned}
\end{equation}

The recent advanced state space model, Mamba~\cite{gu2023mamba}, has further improved 
$\overline{B}$, ${C}$ and $\Delta$ to be input-dependent, thus allowing for a dynamic feature representation.

\subsection{Overall Pipeline}
Given one tamperd image $I\in \mathbb{R}^{H\times W \times 3}$, the objective of the image forgery localization task is to generate the localization map $P \in \mathbb{R}^{H\times W \times 1}$. As illustrated in Fig. \ref{framework}, the overall pipeline of LoMa consists of three main components: global pixel dependency modeling, local pixel dependency modeling, and feature decoding. Firstly, the image is divided into patches to obtain $\hat{I}$. Then, we employ a set of mixed-ssm blocks to extract global features from $\hat{I}$, progressively reducing the resolution to facilitate more efficient global pixel dependency modeling. This process can be formulated as:
\begin{equation} 
X_1, X_2 = \text{Mixed-SSM-Blocks}(\hat{I}) 
\end{equation}
where $X_1 \in \mathbb{R}^{\frac{H}{4} \times \frac{W}{4} \times C_1}$ and $X_2 \in \mathbb{R}^{\frac{H}{8} \times \frac{W}{8} \times C_2}$. Next, we perform local pixel dependency modeling using inverted residual blocks \cite{sandler2018mobilenetv2}:
\begin{equation} 
X_3, X_4 = \text{Inverted-Res-Blocks}(X_2) 
\end{equation}
where $X_3 \in \mathbb{R}^{\frac{H}{16} \times \frac{W}{16} \times C_3}$ and $X_4 \in \mathbb{R}^{\frac{H}{32} \times \frac{W}{32} \times C_4}$. Such allocation of SSM and CNN blocks is aimed at promoting SSMs in the early stages with high resolutions for better global capture, while adopting CNNs at low resolutions for improved efficiency.

Finally, the pixel-wise forgery localization map $P$ is obtained by a simple decoder \cite{xie2021segformer}:
\begin{equation} 
P = \text{Decoder}(X_1, X_2, X_3, X_4) 
\end{equation}

\subsection{Efficient SS2D}
In terms of visual tasks, VMamba \cite{liu2024vmamba} proposed the 2D Selective Scan (SS2D), which maintains the integrity of 2D image structures by scanning four directed feature sequences. Each sequence is processed independently within an S6 block and then combined to form a comprehensive 2D feature map. In this paper, an efficient atrous selective scan \cite{pei2024efficientvmamba} is used. As shown in Fig. \ref{ss2d}, given a feature map with $N$ patches, instead of cross-scanning all patches, we skip scanning patches with a step of 2 and partition them into selected spatial dimensional features. This efficient scan approach reduces computational demands while preserving global feature maps. Drawing inspiration from \cite{guo2025mambair}, who identified the lack of locality and inter-channel interaction in SSMs, we replace the MLP layer with a Channel-Attention Block (CAB) \cite{hu2018squeeze}.

\begin{table*}
\caption{Image forgery localization performance F1[\%] and fIoU[\%]. The best results are highlighted in \textcolor{red}{\textbf{red}}, and the second-best results are highlighted in \textcolor{blue}{\textbf{blue}}.}
\begin{adjustbox}{width=\textwidth}
\begin{tabular}{r|c|cccccccccccccccccccc|cc}
\toprule[1.5pt]
\multirow{2}{*}{Method} &\multirow{2}{*}{Architecture} & \multicolumn{2}{c}{Columbia (160)} & \multicolumn{2}{c}{DSO (100)} & \multicolumn{2}{c}{CASIAv1 (920)} & \multicolumn{2}{c}{NIST (564)} & \multicolumn{2}{c}{Coverage (100)}& \multicolumn{2}{c}{Korus (220)}& \multicolumn{2}{c}{Wild (201)} & \multicolumn{2}{c}{CoCoGlide (512)} & \multicolumn{2}{c}{MISD (227)}  & \multicolumn{2}{c|}{FF++ (1000)} & \multicolumn{2}{c}{\textbf{Average}} \\ \cmidrule{3-24}
                         & & F1  & IoU       & F1    & IoU      & F1   & IoU         & F1   & IoU       & F1   & IoU  & F1 & IoU     & F1    & IoU     & F1     & IoU      & F1     & IoU   & F1     & IoU  & F1     & IoU\\         
\midrule
PSCC \cite{liu2022pscc}  &CNN &61.5 &48.3 &41.1 &31.6 &46.3 &41.0 &18.7 &13.5 &44.4 &33.6  & 10.2 &5.8 &10.8 &8.1 &42.1 &33.2 &65.6 &52.4   &7.0  &4.2 &29.8 &23.9       \\
MVSS++ \cite{dong2022mvss}  &CNN &68.4 &59.6 &27.1 &18.8 &45.1 &39.7 &29.4 &24.0 &44.5 &37.9  &9.5 &6.7 &29.5 &21.9 &35.6 &27.5 &65.9 &52.4   &16.5  &12.7 &33.4 &27.4       \\
CAT-Net \cite{kwon2022learning}    &CNN &79.3 &74.6 &47.9 &40.9 &71.0 &63.7 &30.2 &23.5 &28.9 &23.0  & 6.1 &4.2 &34.1 &28.9 &36.3 &28.8 &39.4 &31.3    &12.3  &9.5 &37.6 &32.0       \\
\midrule 
EITLNet \cite{guo2024effective}   &Transformer   &87.6 &84.2 &42.2 &33.0 &55.7 &52.0 &33.0 &26.7 &44.3 &35.3  &\textcolor{blue}{\textbf{32.3}}&\textcolor{blue}{\textbf{26.1}}&51.9 &43.0 &35.4 &28.8 &\textcolor{red}{\textbf{75.5}} &\textcolor{red}{\textbf{63.8}}   &15.1  &10.7&40.1 &34.3   \\
TruFor \cite{guillaro2023trufor}   &Transformer   &79.8 &74.0 &\textcolor{red}{\textbf{91.0}} &\textcolor{red}{\textbf{86.5}} &69.6 &63.2 &\textcolor{red}{\textbf{47.2}} &\textcolor{red}{\textbf{39.6}} &\textcolor{blue}{\textbf{52.3}} &\textcolor{blue}{\textbf{45.0}}  &\textcolor{red}{\textbf{37.7}}&\textcolor{red}{\textbf{29.9}}&\textcolor{red}{\textbf{61.2}} &\textcolor{red}{\textbf{51.9}} &35.9 &29.1 &60.0 &47.5    &69.2  &\textcolor{blue}{\textbf{56.5}}&59.8 &51.1   \\
MPC \cite{lou2024exploring}   &Transformer   &\textcolor{red}{\textbf{94.5}} &\textcolor{red}{\textbf{93.4}} &\textcolor{blue}{\textbf{51.5}} &\textcolor{blue}{\textbf{39.2}} &\textcolor{blue}{\textbf{74.5}} &\textcolor{blue}{\textbf{68.8}} &43.6 &36.5 &\textcolor{red}{\textbf{61.5}} &\textcolor{red}{\textbf{52.5}}  &29.1&22.2&\textcolor{blue}{\textbf{59.1}} &\textcolor{blue}{\textbf{50.1}} &\textcolor{blue}{\textbf{42.1}} &\textcolor{blue}{\textbf{33.2}} &\textcolor{blue}{\textbf{72.1}} &\textcolor{blue}{\textbf{59.3}}    &\textcolor{blue}{\textbf{69.4}}  &54.9&\textcolor{blue}{\textbf{61.2}} &\textcolor{blue}{\textbf{52.0}}   \\
\midrule 
LoMa (Ours)                      &Mamba &\textcolor{blue}{\textbf{88.9}} &\textcolor{blue}{\textbf{85.2}} &40.7 &29.5 &\textcolor{red}{\textbf{76.6}}&\textcolor{red}{\textbf{69.9}} &\textcolor{blue}{\textbf{45.9}} &\textcolor{blue}{\textbf{38.5}} &52.1 &43.0  & 27.5&21.5&54.2 &44.2 &\textcolor{red}{\textbf{43.2}} &\textcolor{red}{\textbf{33.7}} &68.2 &55.3   &\textcolor{red}{\textbf{71.9}}  &\textcolor{red}{\textbf{60.8}}&\textcolor{red}{\textbf{61.5}} &\textcolor{red}{\textbf{52.7}}         \\

\bottomrule[1.5pt]
\end{tabular}
\end{adjustbox}
\label{shiyan}
\end{table*}

\begin{table}[!t]
\caption{Complexity of Our method compared to SoTA models. The lowest computational complexity is highlighted in \textcolor{red}{\textbf{red}}.}
\tabcolsep=8 pt
\centering
\begin{adjustbox}{width=\linewidth}
\begin{tabular}{r|l|ccc}
\toprule[1.5pt]
\multirow{2}{*}{Method} &\multirow{2}{*}{Year-Venue}  & \multirow{2}{*}{Params. (M)} & \multirow{2}{*}{\begin{tabular}[c]{@{}c@{}}512×512 \\ FLOPs (G)\end{tabular}} & \multirow{2}{*}{\begin{tabular}[c]{@{}c@{}}1024×1024\\ FLOPs (G)\end{tabular}} \\
                        &                              &                                                                               &                                                                                \\
\midrule
PSCC \cite{liu2022pscc} &2022-TCSVT & 3           & 120          & 416               \\
MVSS++ \cite{dong2022mvss}  &2022-TPAMI& 147            & 167          & 683                \\
CAT-Net \cite{kwon2022learning}  &2022-IJCV                & 114            & 134          & 538                \\
\midrule
EITLNet \cite{guo2024effective} &2024-ICASSP & 52            & 83                        & 426                    \\
TruFor \cite{guillaro2023trufor} &2023-CVPR                  & 69             & 231       & 1016               \\
MPC \cite{lou2024exploring}  &2025-TIFS& 45            & 86                        & 335                    \\
\midrule
LoMa (Ours) &2025                 & 37              & \textcolor{red}{\textbf{64}}       & \textcolor{red}{\textbf{258}}               \\
\bottomrule[1.5pt]
\end{tabular}
\end{adjustbox}
\label{flops}
\vspace{-0.2cm}
\end{table}

\section{Experiments}
\label{sec:experiments}

\subsection{Experimental Settings}

\subsubsection{Datasets}
{Consistent with CAT-Net\cite{kwon2022learning}, TruFor \cite{guillaro2023trufor} and MPC \cite{lou2024exploring}, our model is trained on the CAT-Net dataset \cite{kwon2022learning}. Comprehensive evaluation is conducted using 10 test datasets, including Columbia \cite{hsu2006columbia}, DSO \cite{carvalho2015illuminant}, CASIAv1 \cite{dong2013casia}, NIST \cite{guan2019mfc}, Coverage \cite{wen2016coverage},  Korus\cite{korus2016evaluation}, Wild \cite{huh2018fighting}, CocoGlide\cite{guillaro2023trufor}, MISD\cite{kadam2021multiple} and FF++\cite{rossler2019faceforensics++}.}

\subsubsection{Compared Methods}
{
To validate the effectiveness of state space models, we selected 6 representative image forgery localization methods from the most advanced CNN-based and Transformer-based approaches for comparison. The CNN-based methods include PSCC\cite{liu2022pscc}, MVSS++\cite{dong2022mvss}, and CAT-Net\cite{kwon2022learning}, while the Transformer-based methods include EITLNet\cite{guo2024effective}, TruFor\cite{guillaro2023trufor} and MPC \cite{lou2024exploring}.
}

\subsubsection{Evaluation metrics}
{Similar to previous works \cite{guo2024effective, lou2024exploring}, the accuracy of pixel-level forgery localization is measured by F1 and IoU. The fixed threshold 0.5 is adopted to binarize the localization probability map. We follow the approach outlined in \cite{lou2024exploring} to calculate the average F1 and IoU.}

\subsubsection{Implementation details}
{The proposed LoMa is implemented by PyTorch.  We train the network on a single A100 40G GPU. The learning rate starts from 1e-4 and decreases by the plateau strategy. AdamW is adopted as the optimizer, batch size is 8 and all the images used in training are resized to $512 \times 512$ pixels. The common data augmentations, including flipping, blurring, compression and noising are adopted.

\subsubsection{Loss function}
{In line with \cite{guo2024effective}, we utilize a hybrid loss consisting of the DICE \cite{wei2021learn} and Focal \cite{lin2017focal} losses.}

\subsection{Comparison with State-of-the-art Methods}
Table \ref{shiyan} reports the compared results on 10 publicly datasets and Fig. \ref{Visual} visualizes the predicted masks of the methods with publicly available codes. Table \ref{shiyan} demonstrates that LoMa achieves the best performance on 3 test datasets and the second-best performance on 2 test datasets. Additionally, LoMa exhibits the highest average localization accuracy across all ten test datasets. For instance, on the CASIAv1 test dataset, the best CNN-based method, CAT-Net, achieves F1 score 0.710, while the best Transformer-based method, MPC, attains F1 score 0.745. In contrast, LoMa achieves a superior F1 score 0.766. Overall, compared to CNN-based and Transformer-based methods, the proposed LoMa demonstrates superior performance in the image forgery localization task.

We further compare computational complexity of each localization model, as shown in Table \ref{flops}. Due to the lack of global receptive field, CNN-based methods typically require more complex models to capture tampering traces, leading to an increase in the number of parameters. On the other hand, Transformer models exhibit quadratic complexity, significantly increasing computational requirements. In contrast, our LoMa, based on the Mamba architecture, offers both a global receptive field and linear complexity, achieving superior localization performance with lower computational complexity.

In summary, our proposed LoMa not only outperforms existing CNN-based and Transformer-based methods in terms of localization accuracy but also achieves a significant reduction in computational complexity.  The Mamba architecture's ability to maintain efficient processing while ensuring broad contextual understanding makes it particularly well-suited for real-time applications and large-scale datasets, thereby offering a significant advantage over traditional methods in both performance and efficiency.

\begin{figure*}[!t]
\centering
\includegraphics[width=\textwidth]{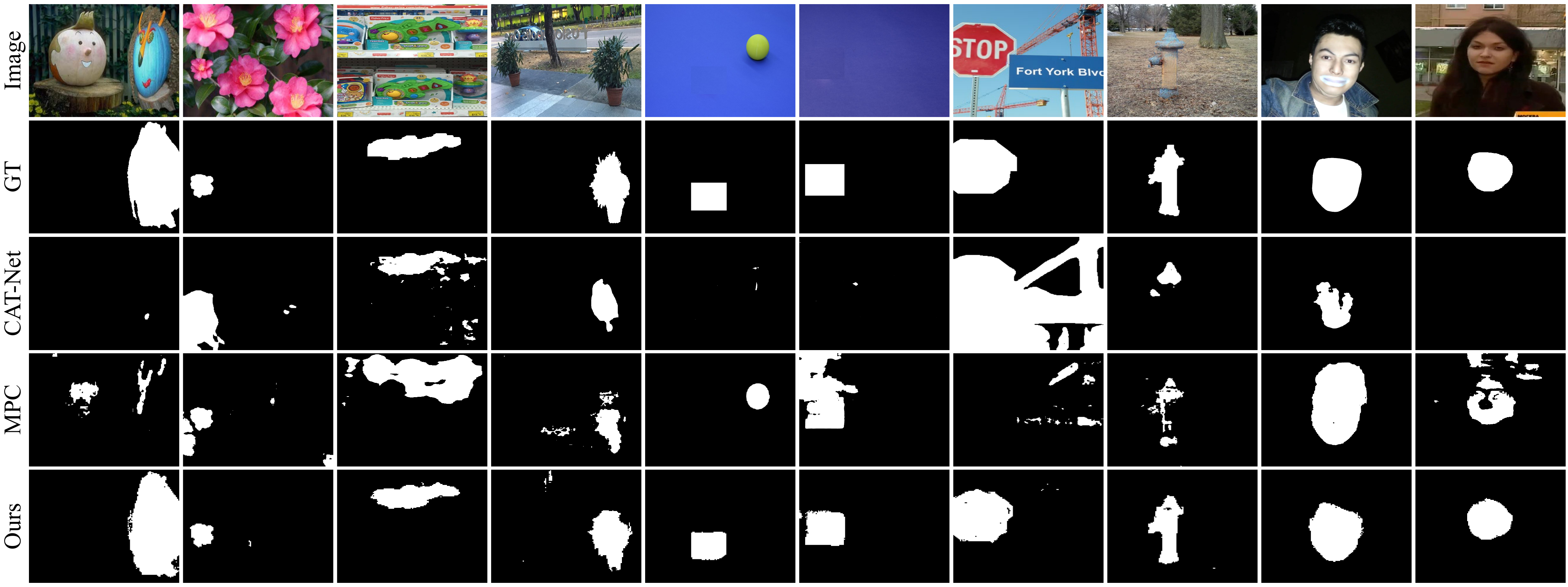}
\caption{Qualitative comparison of forgery localization methods on example testing images. From left to right: two splicing images, two copy-move images, two object-removal images, two AI edited images and two deepfake images. From top to bottom: tampered image, ground truth (GT), and the
localization results from the best CNN-based method CAT-Net, the best Transformer-based method MPC and LoMa.}
\label{Visual}
\end{figure*}

\begin{table}
\caption{Robustness performance F1[\%] and IoU[\%] against online social networks (OSNs) post-processing, including Facebook (FB), Wechat (WC), Weibo (WB) and Whatsapp (WA). The best results are highlighted in \textcolor{red}{\textbf{red}}.}
\centering
\tabcolsep=5 pt
\begin{adjustbox}{width=\linewidth}
\begin{tabular}{r|c|cccccccc|cc}
\toprule[1.5pt]
\multirow{2}{*}{Method} & \multirow{2}{*}{OSNs}     & \multicolumn{2}{c}{CASIAv1} & \multicolumn{2}{c}{Columbia} & \multicolumn{2}{c}{NIST} & \multicolumn{2}{c}{DSO}  & \multicolumn{2}{|c}{Average}\\ \cmidrule{3-12}
                        &                           & F1           & IoU          & F1            & IoU          & F1          & IoU         & F1           & IoU        & F1           & IoU \\         
\midrule                    
CAT-Net \cite{kwon2022learning}                 & \multirow{3}{*}{FB} & 63.3   & 55.9    & 91.8   &90.0   & 15.1   &11.9   & 12.1   &9.8   &47.4 &42.2  \\
MPC \cite{lou2024exploring}                &                     &70.9    &64.4     & \textcolor{red}{\textbf{95.8}}   &\textcolor{red}{\textbf{95.0}}   &42.9    &35.7   & \textcolor{red}{\textbf{52.0}}   &\textcolor{red}{\textbf{39.6}}   &63.1 &56.6\\
LoMa (Ours)                 &                     &\textcolor{red}{\textbf{74.7}}    &\textcolor{red}{\textbf{67.4}}     &88.6   &84.9   &\textcolor{red}{\textbf{45.9}}    &\textcolor{red}{\textbf{38.3}}   &41.3   &30.1    &\textcolor{red}{\textbf{64.7}} &\textcolor{red}{\textbf{57.5}} \\

\midrule
CAT-Net \cite{kwon2022learning}                 & \multirow{3}{*}{WC} & 13.9   & 10.6    & 84.8   &80.8   & 19.1   & 14.9  & 1.7   &1.0   &21.4 &17.9  \\
MPC \cite{lou2024exploring}                &                     & 61.2   & 53.8    & \textcolor{red}{\textbf{94.7}}   &\textcolor{red}{\textbf{93.8}}   & 42.3   & 35.0  & \textcolor{red}{\textbf{50.5}}   &\textcolor{red}{\textbf{38.0}}  &57.6 &50.6            \\
LoMa (Ours)                &                     &\textcolor{red}{\textbf{64.2}}    &\textcolor{red}{\textbf{56.0}}     &89.0   &85.3   &\textcolor{red}{\textbf{45.7}}    &\textcolor{red}{\textbf{38.0}}   &43.0    &31.5  &\textcolor{red}{\textbf{59.3}} &\textcolor{red}{\textbf{51.5}}   \\

\midrule
CAT-Net \cite{kwon2022learning}                 & \multirow{3}{*}{WB} &42.5    &36.2     & 92.1   & 89.7  & 20.8   & 16.0  & 2.3  &1.3  &37.7 &32.6    \\
MPC \cite{lou2024exploring}                &                     &70.7    &64.7     & \textcolor{red}{\textbf{94.8}}   & \textcolor{red}{\textbf{94.2}}  & 43.3   & 36.1  & \textcolor{red}{\textbf{51.9}}   &\textcolor{red}{\textbf{39.8}}  &63.0 &56.7   \\
LoMa (Ours)                &                     &\textcolor{red}{\textbf{73.5}}    &\textcolor{red}{\textbf{66.6}}     &88.7   &85.0   &\textcolor{red}{\textbf{46.4}}    &\textcolor{red}{\textbf{39.0}}   &43.6    &32.0  &\textcolor{red}{\textbf{64.4}} &\textcolor{red}{\textbf{57.4}}   \\

\midrule
CAT-Net \cite{kwon2022learning}                 & \multirow{3}{*}{WA} & 42.3   & 37.8    & 92.1   &89.9   & 20.1   & 16.8  & 2.2   &1.5  &37.4 &33.7            \\
MPC \cite{lou2024exploring}                &                     & 69.9   & 63.5    & \textcolor{red}{\textbf{95.0}}   &\textcolor{red}{\textbf{94.1}}   & 43.8   & 36.3  & \textcolor{red}{\textbf{53.4}}   &\textcolor{red}{\textbf{41.0}}   &62.9 &56.2              \\
LoMa  (Ours)               &                     &\textcolor{red}{\textbf{73.7}}    &\textcolor{red}{\textbf{66.4}}     &88.9   &85.2   &\textcolor{red}{\textbf{47.0}}    &\textcolor{red}{\textbf{39.8}}   &43.0    &31.4 &\textcolor{red}{\textbf{64.7}}&\textcolor{red}{\textbf{57.5}}    \\

\bottomrule[1.5pt]
\end{tabular}
\end{adjustbox}
\label{osn}
\end{table}

\begin{table}[!t]
\caption{Comparison of localization performance for ablation studies. Metric values are in percentage. The best results are highlighted in \textcolor{red}{\textbf{red}}.}
\tabcolsep=10 pt
\centering
\begin{adjustbox}{width=\linewidth}
\begin{tabular}{l|ccccccccccccc}
\toprule[1.5pt]
\multirow{2}{*}{Method}  & \multicolumn{2}{c}{Columbia} & \multicolumn{2}{c}{CASIAv1} & \multicolumn{2}{c}{FF++}\\ \cmidrule{2-7}
  & F1            & IoU          & F1         & IoU        & F1           & IoU            \\ \midrule
 w/o Focal Loss                 & 79.4         & 75.3        & 66.4      & 59.8      & 63.4        & 51.9               \\
 w/o DICE Loss              &  72.7         &  69.5        & 60.8      & 54.4      & 58.5        & 48.2               \\
 w/o CAB              &  85.9         &  82.6        & 74.7      & 66.3      & 68.8        & 57.8          \\
LoMa (Ours)          & \textcolor{red}{\textbf{88.9}}        & \textcolor{red}{\textbf{85.2}}        & \textcolor{red}{\textbf{76.6}}      & \textcolor{red}{\textbf{69.9}}      & \textcolor{red}{\textbf{71.9}}       & \textcolor{red}{\textbf{60.8}} \\
\bottomrule[1.5pt]
\end{tabular}
\end{adjustbox}
\label{ablation}
\end{table}

\subsection{Robustness Evaluation}

We first assess the robustness of image forgery localization methods against the complex post operations introduced by online social networks (OSNs).  Following the prior work \cite{wu2022robust}, the four forgery datasets transmitted through Facebook, Weibo, Wechat and Whatsapp platforms are tested. Table \ref{osn} shows that, our method mostly achieves the highest accuracy across the four datasets for each social network platform. Among the compared methods, LoMa enjoys the smallest performance loss incurred by OSNs. Note that CAT-Net achieves higher performance on the processed Columbia dataset due to its specialized learning of JPEG compression artifacts. However, it shows obvious performance decline on other datasets, for instance, with F1=0.918 on the Facebook version of Columbia while with F1=0.139 on the Wechat version of CASIAv1. In contrast, our LoMa consistently keeps high localization accuracy across all OSN transmissions. Such results verify the robustness of LoMa against online social networks.

\subsection{Ablation Studies}
We conduct extensive ablation studies to validate the effectiveness of LoMa, as shown in Table \ref{ablation}. First, we drop the Focal or DICE loss in the hybrid loss function. We can observe that the performance of our model deteriorates if either loss function is missing. It suggests that such two loss functions both play important roles in optimizing the LoMa. Then, if the CAB is removed from the Mixed-SSM block and replaced with a native MLP, a slight decline in the localization performance of LoMa can be observed. Such a result indicates that enhancing channel attention can further unlock the potential of Mamba architecture.

\section{Conclusion}
\label{sec:conclusion}

In this work, we propose LoMa, which is the first image forgery localization method based on state space models. LoMa employs a selective state space model to capture global pixel dependencies. Compared to CNN-based methods, LoMa has a global receptive field, and compared to Transformer-based methods, LoMa exhibits linear computational complexity. Extensive evaluation results have verified the effectiveness of our scheme and demonstrated the great potential of state space models in forensic tasks. In the future, we will explore more efficient applications of state space models in the field of digital forensics.

\bibliographystyle{IEEEtran}
\bibliography{ref}

\begin{thebibliography}{10}
\providecommand{\url}[1]{#1}
\csname url@samestyle\endcsname
\providecommand{\newblock}{\relax}
\providecommand{\bibinfo}[2]{#2}
\providecommand{\BIBentrySTDinterwordspacing}{\spaceskip=0pt\relax}
\providecommand{\BIBentryALTinterwordstretchfactor}{4}
\providecommand{\BIBentryALTinterwordspacing}{\spaceskip=\fontdimen2\font plus
\BIBentryALTinterwordstretchfactor\fontdimen3\font minus \fontdimen4\font\relax}
\providecommand{\BIBforeignlanguage}[2]{{%
\expandafter\ifx\csname l@#1\endcsname\relax
\typeout{** WARNING: IEEEtran.bst: No hyphenation pattern has been}%
\typeout{** loaded for the language `#1'. Using the pattern for}%
\typeout{** the default language instead.}%
\else
\language=\csname l@#1\endcsname
\fi
#2}}
\providecommand{\BIBdecl}{\relax}
\BIBdecl

\bibitem{liu2022pscc}
X.~Liu, Y.~Liu, J.~Chen, and X.~Liu, ``{PSCC-Net}: Progressive spatio-channel correlation network for image manipulation detection and localization,'' \emph{IEEE Transactions on Circuits and Systems for Video Technology}, vol.~32, no.~11, pp. 7505--7517, 2022.

\bibitem{dong2022mvss}
C.~Dong, X.~Chen, R.~Hu, J.~Cao, and X.~Li, ``{MVSS-Net}: Multi-view multi-scale supervised networks for image manipulation detection,'' \emph{IEEE Transactions on Pattern Analysis and Machine Intelligence}, vol.~45, no.~3, pp. 3539--3553, 2022.

\bibitem{kwon2022learning}
M.-J. Kwon, S.-H. Nam, I.-J. Yu, H.-K. Lee, and C.~Kim, ``Learning jpeg compression artifacts for image manipulation detection and localization,'' \emph{International Journal of Computer Vision}, vol. 130, no.~8, pp. 1875--1895, 2022.

\bibitem{guo2024effective}
K.~Guo, H.~Zhu, and G.~Cao, ``Effective image tampering localization via enhanced transformer and co-attention fusion,'' in \emph{Proceedings of the IEEE International Conference on Acoustics, Speech and Signal Processing}, 2024, pp. 4895--4899.

\bibitem{guillaro2023trufor}
F.~Guillaro, D.~Cozzolino, A.~Sud, N.~Dufour, and L.~Verdoliva, ``{TruFor}: Leveraging all-round clues for trustworthy image forgery detection and localization,'' in \emph{Proceedings of the IEEE/CVF Conference on Computer Vision and Pattern Recognition}, 2023, pp. 20\,606--20\,615.

\bibitem{lou2024exploring}
Z.~Lou, G.~Cao, K.~Guo, H.~Zhu, and L.~Yu, ``Exploring multi-view pixel contrast for general and robust image forgery localization,'' \emph{IEEE Transactions on Information Forensics and Security}, 2025.

\bibitem{gu2021efficiently}
A.~Gu, K.~Goel, and C.~R{\'e}, ``Efficiently modeling long sequences with structured state spaces,'' in \emph{Proceedings of the International Conference on Learning Representations}, 2021.

\bibitem{gu2023mamba}
A.~Gu and T.~Dao, ``Mamba: Linear-time sequence modeling with selective state spaces,'' \emph{arXiv preprint arXiv:2312.00752}, 2023.

\bibitem{sandler2018mobilenetv2}
M.~Sandler, A.~Howard, M.~Zhu, A.~Zhmoginov, and L.-C. Chen, ``{MobileNetV2}: Inverted residuals and linear bottlenecks,'' in \emph{Proceedings of the IEEE/CVF Conference on Computer Vision and Pattern Recognition}, 2018, pp. 4510--4520.

\bibitem{xie2021segformer}
E.~Xie, W.~Wang, Z.~Yu, A.~Anandkumar, J.~M. Alvarez, and P.~Luo, ``{SegFormer}: Simple and efficient design for semantic segmentation with transformers,'' \emph{Advances in Neural Information Processing Systems}, vol.~34, pp. 12\,077--12\,090, 2021.

\bibitem{liu2024vmamba}
Y.~Liu, Y.~Tian, Y.~Zhao, H.~Yu, L.~Xie, Y.~Wang, Q.~Ye, and Y.~Liu, ``{VMamba}: Visual state space model,'' \emph{arXiv preprint arXiv:2401.10166}, 2024.

\bibitem{pei2024efficientvmamba}
X.~Pei, T.~Huang, and C.~Xu, ``{EfficientVMamba}: Atrous selective scan for light weight visual mamba,'' \emph{arXiv preprint arXiv:2403.09977}, 2024.

\bibitem{guo2025mambair}
H.~Guo, J.~Li, T.~Dai, Z.~Ouyang, X.~Ren, and S.-T. Xia, ``{MambaIR}: A simple baseline for image restoration with state-space model,'' in \emph{Proceedings of the Conference on European Conference on Computer Vision}, 2025, pp. 222--241.

\bibitem{hu2018squeeze}
J.~Hu, L.~Shen, and G.~Sun, ``Squeeze-and-excitation networks,'' in \emph{Proceedings of the IEEE/CVF Conference on Computer Vision and Pattern Recognition}, 2018, pp. 7132--7141.

\bibitem{hsu2006columbia}
J.~Hsu and S.~Chang, ``Columbia uncompressed image splicing detection evaluation dataset,'' \emph{Columbia DVMM Research Lab}, 2006.

\bibitem{carvalho2015illuminant}
T.~Carvalho, F.~A. Faria, H.~Pedrini, R.~d.~S. Torres, and A.~Rocha, ``Illuminant-based transformed spaces for image forensics,'' \emph{IEEE Transactions on Information Forensics and Security}, vol.~11, no.~4, pp. 720--733, 2015.

\bibitem{dong2013casia}
J.~Dong, W.~Wang, and T.~Tan, ``{CASIA} image tampering detection evaluation database,'' in \emph{Proceedings of the IEEE China Summit and International Conference on Signal and Information Processing}, 2013, pp. 422--426.

\bibitem{guan2019mfc}
H.~Guan, M.~Kozak, E.~Robertson, Y.~Lee, A.~N. Yates, A.~Delgado, D.~Zhou, T.~Kheyrkhah, J.~Smith, and J.~Fiscus, ``{MFC Datasets}: Large-scale benchmark datasets for media forensic challenge evaluation,'' in \emph{Proceedings of the IEEE Winter Applications of Computer Vision Workshops}, 2019, pp. 63--72.

\bibitem{wen2016coverage}
B.~Wen, Y.~Zhu, R.~Subramanian, T.-T. Ng, X.~Shen, and S.~Winkler, ``{COVERAGE}—{A} novel database for copy-move forgery detection,'' in \emph{Proceedings of the IEEE International Conference on Image Processing}, 2016, pp. 161--165.

\bibitem{korus2016evaluation}
P.~Korus and J.~Huang, ``Evaluation of random field models in multi-modal unsupervised tampering localization,'' in \emph{Proceedings of the IEEE International Workshop on Information Forensics and Security}, 2016, pp. 1--6.

\bibitem{huh2018fighting}
M.~Huh, A.~Liu, A.~Owens, and A.~A. Efros, ``{Fighting Fake News}: Image splice detection via learned self-consistency,'' in \emph{Proceedings of the European Conference on Computer Vision}, 2018, pp. 101--117.

\bibitem{kadam2021multiple}
K.~D. Kadam, S.~Ahirrao, and K.~Kotecha, ``{Multiple Image Splicing Dataset (MISD)}: A dataset for multiple splicing,'' \emph{Data}, vol.~6, no.~10, p. 102, 2021.

\bibitem{rossler2019faceforensics++}
A.~Rossler, D.~Cozzolino, L.~Verdoliva, C.~Riess, J.~Thies, and M.~Nie{\ss}ner, ``{FaceForensics++}: Learning to detect manipulated facial images,'' in \emph{Proceedings of the IEEE/CVF International Conference on Computer Vision}, 2019, pp. 1--11.

\bibitem{wei2021learn}
Q.~Wei, X.~Li, W.~Yu, X.~Zhang, Y.~Zhang, B.~Hu, B.~Mo, D.~Gong, N.~Chen, D.~Ding \emph{et~al.}, ``Learn to segment retinal lesions and beyond,'' in \emph{Proceedings of the IEEE International Conference on Pattern Recognition}, 2021, pp. 7403--7410.

\bibitem{lin2017focal}
T.-Y. Lin, P.~Goyal, R.~Girshick, K.~He, and P.~Doll{\'a}r, ``Focal loss for dense object detection,'' in \emph{Proceedings of the IEEE/CVF International Conference on Computer Vision}, 2017, pp. 2980--2988.

\bibitem{wu2022robust}
H.~Wu, J.~Zhou, J.~Tian, J.~Liu, and Y.~Qiao, ``Robust image forgery detection against transmission over online social networks,'' \emph{IEEE Transactions on Information Forensics and Security}, vol.~17, pp. 443--456, 2022.

\end{thebibliography}

\end{document}